\documentclass{article}

\usepackage[final]{neurips_2022}

\usepackage[utf8]{inputenc} 
\usepackage[T1]{fontenc}    
\usepackage[hidelinks]{hyperref}       
\usepackage{url}            
\usepackage{booktabs}       
\usepackage{amsfonts}       
\usepackage{nicefrac}       
\usepackage{microtype}      
\usepackage{xcolor}         

\usepackage{amsmath}
\usepackage{amssymb}
\usepackage{graphicx}
\usepackage{algorithm}
\usepackage{algorithmic}
\usepackage{todonotes}
\usepackage{amsthm}
\usepackage[noabbrev,capitalise]{cleveref}
\usepackage{thmtools}
\usepackage{thm-restate}
\usepackage{xspace}
\newcommand{\FlowSelect}{\textsc{FlowSelect}\xspace}
\newcommand{\FlowSelectNS}{\textsc{FlowSelect}}
\DeclareMathOperator*{\argmax}{arg\,max}
\DeclareMathOperator*{\argmin}{arg\,min}

\title{Normalizing Flows for Knockoff-free Controlled Feature Selection}

\author{%
  Derek Hansen\\
  Department of Statistics\\
  University of Michigan\\
  \texttt{dereklh@umich.edu} \\
  \And
  Brian Manzo \\
  Department of Statistics\\
  University of Michigan\\
  \texttt{bmanzo@umich.edu} \\
  \And
  Jeffrey Regier \\
  Department of Statistics\\
  University of Michigan\\
  \texttt{regier@umich.edu} \\
}

\begin{document}

\maketitle

\begin{abstract}
Controlled feature selection aims to discover the features a response depends on while limiting the false discovery rate (FDR) to a predefined level.
Recently, multiple deep-learning-based methods have been proposed to perform controlled feature selection through the Model-X knockoff framework.
We demonstrate, however, that these methods often fail to control the FDR for two reasons.
First, these methods often learn inaccurate models of features. Second, the ``swap'' property, which is required for knockoffs to be valid, is often not well enforced.
We propose a new procedure called \FlowSelect
to perform controlled feature selection that does not suffer from either of these two problems.
To more accurately model the features, \FlowSelect uses normalizing flows, the state-of-the-art method for density estimation.
Instead of enforcing the ``swap'' property, \FlowSelect uses a novel MCMC-based procedure to calculate p-values for each feature directly.
Asymptotically,
\FlowSelect computes valid p-values.
Empirically, \FlowSelect consistently controls the FDR on both synthetic and semi-synthetic benchmarks, whereas competing knockoff-based approaches do not.
\FlowSelect also demonstrates greater power on these benchmarks.
Additionally, \FlowSelect correctly infers the genetic variants associated with specific soybean traits from GWAS data.
\end{abstract}

\section{Introduction}
Researchers in machine learning have made much progress in developing regression and classification models that can predict a response based on features.
In many application areas, however, practitioners need to know \textit{which} features drive variation in the response, and they need to do so in a way that limits the number of false discoveries.
For example, in genome-wide association studies (GWAS), scientists must consider
hundreds of thousands of genetic markers to identify variants associated with a particular trait or disease.
The cost of false discoveries (i.e., selecting variants that are not associated with the disease) is high, as a costly follow-up experiment is often conducted for each selected variant.
Another example where controlled feature selection matters
is analyzing observational data about the effectiveness of educational interventions. In this case, researchers may want to select certain educational programs to implement on a larger scale and require confidence that their selection does not include unacceptably many ineffective programs.
As a result, researchers are interested in methods that model the dependence structure of the data while providing an upper bound on the false discovery rate (FDR).

Model-X knockoffs \citep{candesPanningGoldModelX2018} is a popular method for controlled variable selection, offering theoretical guarantees of FDR control and the flexibility to use arbitrary predictive models.
However, even with knowledge of the underlying feature distribution, the Model-X knockoffs method is not feasible unless the feature distribution is either a finite mixture of Gaussians \citep{giminezKnockoffsMassNew2019} or has a known Markov structure \citep{batesMetropolizedKnockoffSampling2020}.
Hence, a body of research explores the use of empirical approaches that use deep generative models to estimate the distribution of $X$ and sample knockoff features \citep{jordon2018knockoffgan, liuAutoEncodingKnockoffGenerator2018, romanoDeepKnockoffs2018, sudarshanDeepDirectLikelihood2020}.

The ability of these methods to control the FDR is contingent on their ability to correctly model the distribution of the features.
By itself, learning a sufficiently expressive feature model can be challenging.
However, the knockoff procedure requires learning a knockoff distribution that satisfies the \textit{swap property}, which is a much stronger requirement.
Formally, let $X \in \mathbb R^D$ be a sample from the feature distribution and $\tilde X \in \mathbb R^D$ be a sample from the knockoff distribution conditioned on $X$.
The swap property stipulates that the joint distribution $(X, \tilde X) \in \mathbb{R}^{2D}$ must be invariant to swapping the positions of any subset of features $S \in \{1, \dots, D\}$:
\begin{equation}
    (X, \tilde X)_{\text{swap}(S)} \overset D = (X, \tilde X)
\end{equation}
Here, $\text{swap}(S)$ means exchanging the positions of $X_j$ and $\tilde X_j$ for all $j \in S$.
For example, in the case $D=3$ and $S=\{1, 3\}$, the joint distribution is $(X, \tilde X) = (X_1, X_2, X_3, \tilde X_1, \tilde X_2, \tilde X_3)$, and the swapped joint distribution is $(X, \tilde X)_{\text{swap}(S)} = (\tilde X_1, X_2, \tilde X_3, X_1, \tilde X_2, X_3)$.
Note that, for $S=\{1, \dots, D\}$, the swap property implies that $\tilde X \overset D = X$.
See \citet{candesPanningGoldModelX2018} for a more detailed description of the swap property.

Even if a distribution were found satisfying the swap property, it may not provide enough power to make discoveries.
For example, both properties are trivially satisfied by constructing exact copies of the features as knockoffs,
but the resulting procedure has no power.

In situations where a valid knockoff distribution is available to sample from, knockoffs are computationally appealing because they require only one sample from a knockoff distribution to assess the relevance of all $p$ features.
However, in situations where the joint density of the features is unknown, we show that empirical approaches to knockoff generation \citep{jordon2018knockoffgan, liuAutoEncodingKnockoffGenerator2018, romanoDeepKnockoffs2018, sudarshanDeepDirectLikelihood2020}
fail to characterize a valid knockoff distribution and therefore do not control the FDR.
We further show that even with a known covariate model, it is not straightforward to construct a valid knockoff distribution unless a specific model structure is known.

We propose a new feature selection method called \FlowSelect (\cref{sec:method}), which does not suffer from these problems.
\FlowSelect uses normalizing flows to learn the joint density of the covariates.
Normalizing flows is a state-of-the-art method for density estimation; asymptotically, it can approximate any distribution arbitrarily well \citep{papamakariosNormalizingFlowsProbabilistic2019a, kobyzevNormalizingFlowsIntroduction2020, huangNeuralAutoregressiveFlows2018a}.
Additionally, \FlowSelect circumvents the need to sample a knockoff distribution by instead applying a fast variant of the conditional randomization test (CRT) introduced in \citet{candesPanningGoldModelX2018}.
Samples from the complete conditionals are drawn using MCMC, ensuring they are unbiased with respect to the learned data distribution.

Asymptotically, \FlowSelect computes correct p-values to use for feature selection (\cref{sec:asymptotic}).
Our proof assumes the universal approximation property of normalizing flows and the convergence of MCMC samples to the Markov chain's stationary distribution.
Under the same assumptions as the CRT, which includes a multiple-testing correction as in \citet{Benjamini1995}, a selection threshold can be picked which controls the FDR at a pre-defined level.
Empirically, on both synthetic (Gaussian) data and semi-synthetic data (real predictors and a synthetic response), \FlowSelect controls the FDR where other deep-learning-based knockoff methods do not.
In cases in which competing methods do control the FDR, \FlowSelect shows higher power (\cref{sec:experiments}).
Finally, in a challenging real-world problem with soybean genome-wide association study (GWAS) data, \FlowSelect successfully harnesses normalizing flows for modeling discrete and sequential GWAS data, and for selecting genetic variants the traits depend on (\cref{sec:soybean}).

\section{Background}\label{sec:background}

\FlowSelect brings together four existing lines of research, which we briefly introduce below.

\paragraph{Normalizing flows} Normalizing flows is a general framework for density estimation of a multi-dimensional distribution with arbitrary dependencies \citep{papamakariosNormalizingFlowsProbabilistic2019a}.
A normalizing flow starts with a simple probability distribution (e.g., Gaussian or uniform), which is called the \textit{base distribution} and denoted $Z$, and transforms samples from this base distribution through a series of invertible and differentiable transformations, denoted $G$, to define the joint distribution of $X \in \mathbb{R}^D \sim \mathcal P_X$.
A normalizing flow with enough transformations can approximate any multivariate density, subject to regularity conditions detailed by \citet{kobyzevNormalizingFlowsIntroduction2020}.
Compared to other density-estimation methods, normalizing flows are computationally efficient.
Details about the specific normalizing flow architecture used in \FlowSelect are provided in Appendix~\ref{sec:flows}.

\paragraph{Controlled feature selection} Consider a response $Y$ which depends on a vector of features $X \in \mathbb{R}^D$.
Depending on how the features are chosen, it is plausible that only a subset of the features contains all relevant information about $Y$.
Specifically, conditioned on the relevant features in $X$, $Y$ is independent of the remaining features in $X$ (i.e. the null features).
The goal of the controlled feature selection procedure is to maximize the number of relevant features selected while limiting the number of null features selected to a predefined level.
If we denote the total number of selected features $R$, then we can decompose $R$ into $V$, the number of relevant features selected, and $S$, the number of null features selected.

\paragraph{Conditional randomization test}
Controlled feature selection can be seen as a multiple hypothesis testing problem where there are $p$ null hypotheses, each of which says that feature $X_j$ is conditionally independent of the response $Y$ given all the other features $X_{-j}$.
Explicitly, the test of the following hypothesis is conducted for each feature $j = \{1, \dots, D\}$:
\begin{equation}\label{eq:hypothesis}
    H_0: X_j \perp Y \vert X_{-j} \quad \text{versus} \quad H_1: X_j \not\perp Y \vert X_{-j}.
\end{equation}
To test these hypotheses, one can use a conditional randomization test (CRT) \citep{candesPanningGoldModelX2018}.
For each feature tested in a conditional randomization test, a test statistic $T_j$ (e.g., the LASSO coefficient or another measure of feature importance) is first computed on the data.
Then, the null distribution of $T_j$ is estimated by computing its value $\tilde T_j$ based on samples $\tilde X_j$ drawn from the conditional distribution of $X_j$ given $X_{-j}$.
Finally, the p-value is calculated based on the empirical CDF of the null test statistics, and features whose p-values fall below the threshold set by the Benjamini-Hochberg procedure \citep{Benjamini1995} are selected.
Though the CRT is introduced as a computationally inefficient alternative to knockoffs, the CRT nonetheless has appeal because it requires only knowledge of the feature distribution, which can be learned empirically by maximum likelihood.

\paragraph{Holdout randomization test}
\label{sec:hrt}
The holdout randomization test (HRT) \citep{tanseyHoldoutRandomizationTest2019} is a fast variant of the CRT; it uses a test statistic that requires fitting the model only once.
Let $\theta$ represent the parameters of the chosen model, and let $T(X, Y, \theta)$ be an importance statistic calculated from the model with input data.
For example, $T$, could be the predictive likelihood $\mathcal P_\theta (Y^{\text{test}}|X^{\text{test}})$ or the predictive score $R^2$.
To use the HRT, first fit model parameters $\hat \theta$ based on the training data.
Next, for each covariate $j$, calculate the test statistic $T^*_j \leftarrow T(X^{\text{test}}, Y^{\text{test}}, \hat \theta)$.
Then, generate $k$ null samples and compute $T_{j, k} \leftarrow T(X^{\text{test}}_{(j \leftarrow j_k)}, Y^{\text{test}}, \hat \theta)$, where $X^{\text{test}}_{(j \leftarrow j_k)}$ replaces the $j$-th covariate with the $k$-th generated null sample.
Finally, calculate the p-value as in the CRT, based on the empirical CDF of the null test statistics.

\section{Methodology}\label{sec:method}
\begin{algorithm}[b]
\caption{Step 2 of the \FlowSelect procedure for drawing $K$ null features $\tilde X_{i, j} | X_{i, -j}$ for feature $j$ at observation $i$.
}\label{alg:mcmc}
\begin{algorithmic}
  \STATE {\textbf{Input:}} {
    Feature matrix $X \in \mathbb R^{N \times D}$, observation index $i$, feature index $j$, number of samples $K$, fitted normalizing flow $p_{\hat \theta}$, MCMC proposal $q_j$
  }
  \STATE {\textbf{Output:}}{
  Null features $\tilde X_{i, j, k}$ for $k=1, \dots, K$
  }
    \FOR{$k = 1, \dots, K$ }
    \STATE Propose: $X_{i, j, k}^\star \sim q_j(\cdot | \tilde X_{i, j, k-1}, X_{i, -j})$
    \STATE $r_{i, j, k} \leftarrow \frac{p_{\hat \theta}(X_{i, j, k}^\star, X_{i, -j}) q_j(\tilde X_{i, j, k-1} | X_{i, j, k}^\star, X_{i, -j})}{p_{\hat \theta}(\tilde X_{i, j, k-1}, X_{i, -j}) q_j(X_{i, j, k}^\star| \tilde X_{i, j, k-1}, X_{i, -j})}$
    \STATE Sample: $U_{i, j, k} \sim \text{Bernoulli} (r_{i, j, k} \wedge 1)$
    \IF{$U_{i,j,k} = 1$}
    \STATE $\tilde X_{i,j,k} \leftarrow X_{i,j,k}^\star$
    \ELSE
    \STATE $\tilde X_{i,j,k} \leftarrow \tilde X_{i,j,k-1}$
    \ENDIF
  \ENDFOR
\end{algorithmic}
\end{algorithm}

\FlowSelect implements the CRT for arbitrary feature distributions by using a normalizing flow to fit the feature distribution and Markov chain Monte Carlo (MCMC) to sample from each complete conditional distribution.
Performing controlled feature selection with \FlowSelect consists of the three steps below.

\subsection*{Step 1: Model the predictors with a normalizing flow}
Starting with the observed samples of the features $X_{1}, \dots, X_{N} \sim \mathcal P_{X}$, we fit the parameters of a normalizing flow $G_\theta$ to maximize the log likelihood of the data with respect to a base distribution $p_Z$:
\begin{equation}
  \label{eq:objective_normalizing_flow}
  \begin{split}
        \hat \theta &= \argmax_\theta \sum_{i=1}^{N} \log p_{\theta}(X_{i}) \\
        \text{where}~p_\theta(X_i) &= p_Z(G_\theta(X)) \left| \det \left( \frac{\partial G_\theta(X)}{\partial X} \right) \right |.
  \end{split}
\end{equation}
The resulting density $p_{\hat \theta}$ is a fitted approximation to the true density $\mathcal P_X$.
The specific normalizing flow architecture we use in our first two experiments consists of a single Gaussianization layer \citep{mengGaussianizationFlows2020b} followed by a masked autoregressive flow (MAF) \citep{papamakariosMaskedAutoregressiveFlow2017}.
The first layer can learn complex marginal distributions for each covariate, while the MAF learns the dependencies between them.
More detail on normalizing flows and on this particular architecture can be found in \cref{sec:flows}.

\subsection*{Step 2: Sample from the complete conditionals with MCMC}

For each feature $j$, we aim to sample corresponding null features $\tilde X_{i, j, k}$ for all $k \in \{1, \dots, K\}$ that are equal in distribution to $p_{\hat \theta}(X_{i, j} | X_{i, -j})$, but independent of $Y_{i}$.
However, directly sampling from this conditional distribution is intractable.
Instead, we implement an MCMC algorithm that admits it as a stationary distribution.
The samples drawn from MCMC are autocorrelated, but any statistic calculated over these samples will converge almost surely to the correct value.
The choice of the MCMC proposal distribution $q_{j}$ is flexible.
Because each Markov chain is only one-dimensional, a Metropolis-Hastings Gaussian random walk with the standard deviation set based on the covariance can be expected to mix rapidly.
Alternatively, information from $p_{\hat \theta}$, such as higher-order derivatives, could be used to construct a more efficient proposal.
\cref{alg:mcmc} details how to implement step 2.

\subsection*{Step 3: Test for significance with the HRT}
As in the CRT, feature $j$ has high evidence of being significant if,
under the assumption that $j$ is a null feature,
the probability of realizing a test statistic greater than the observed $T_j(X)$ is low.
Formally, letting \([\tilde X_j, X_{-j}]\) be the observed feature matrix with the observed feature $X_j$ swapped out with the null feature $\tilde X_j$, we can write this as a p-value $\alpha_j$:
\begin{equation} \label{eq:p_value}
\begin{split}
      \alpha_{j} &\equiv \mathcal P_{\tilde X_j | X_{-j}} \left (T_{j}(X) < T_{j} ([\tilde X_j, X_{-j}])\right).
\end{split}
\end{equation}
However, the above p-value $\alpha_j$ is not tractable.
For each sample $\tilde X_{\cdot, j, k}$ drawn using MCMC, we calculate the corresponding feature statistic
and compare it to the real feature statistic, leading to an approximated p-value $\hat \alpha_j$:
\begin{equation}
    \label{eq:empirical_p_value}
    \hat \alpha_{j} \equiv \frac 1 {K+1} ( 1 + \sum_{k=1}^K \mathbf 1[T_j(X) < T_{j} ([\tilde X_{j, k}, X_{-j}])).
  \end{equation}
  To control the FDR, we use the Benjamini-Hochberg procedure to establish a threshold for the observed p-values.
  Specifically, we set the threshold to
\(
s(\gamma) \triangleq \max_j \{\hat \alpha_{j} : \hat \alpha_j \le \frac{j}{D} \gamma \},
\)
and select all features $j$ such that $\alpha_j \leq s(\gamma)$.

The Benjamini-Hochberg correction only guarantees FDR control provided that the p-values have either positive or zero correlation.
Thus, the FDR control of \FlowSelect depends on these assumptions being met.
A more conservative correction from \citet{benjaminiControlFalseDiscovery2001} allows for arbitrary dependencies in p-values, but it suffers from low power.
The Benjamini-Hochberg correction is widely used and empirically robust \citep{tanseyHoldoutRandomizationTest2019}, so we report results using it.
Across our synthetic and semi-synthetic benchmarks in \cref{sec:experiments}, we also find that \FlowSelect maintains empirical FDR control.

Provided that the Benjamini-Hochberg assumptions are met, the FDR will be controlled, but the power of the test depends on $T_j$ being higher when $j$ is a significant feature.
For example, if $Y$ is expected to vary approximately linearly with respect to $X$, $T_{j}(X)$ could be the absolute estimated regression coefficient $|\hat \beta_{j}|$ for the linear model $Y = X\beta + \epsilon$.
Another choice is the HRT feature statistic described earlier.

\section{Asymptotic results}
\label{sec:asymptotic}

The ability of \FlowSelect to control the FDR relies on its ability to produce estimated p-values that converge to the correct p-values for the hypothesis test in \cref{eq:hypothesis}.

  \begin{restatable}{theorem}{asymptotic}\label{thm:asymptotic}
\begin{samepage}
  Let $X \in \mathbb R_{N \times D}$ be a random feature matrix, where each row $X_{i, \cdot}$ is independent and identically distributed; $x \in \mathbb R_{N \times D}$ be the observed feature matrix;
  and $\alpha_j$ be the p-value as defined in \cref{eq:p_value} with test statistic $T_j(X)$.
  Suppose there exists a sequence of functions $\left( G^{n} \right)_{n = 1}^\infty$ and a base random variable $Z$ satisfying the following conditions:
  \begin{enumerate}
    \item Each $G^n$ is continuously differentiable and invertible.
    \item $G^n \to G$ pointwise for some map $G$ that is triangular, increasing, continuously differentiable, and satisfies $G(X_{i, \cdot}) \overset D = Z$.
\end{enumerate}
For $n=1, 2, \dots$, let $X^n$ be the random feature matrix where each row $i$ is independent and has distribution $X^n_{i, \cdot} = (G^n)^{-1} (Z)$.
Then, the p-value in \cref{eq:empirical_p_value} calculated using $K$ MCMC samples targeting $X_{\cdot, j}^n \mid X_{\cdot, -j}^n = x_{\cdot, -j}$ converges to the correct p-value $\alpha_j$ with probability $1$.
\end{samepage}
\end{restatable}
Here we sketch the proof. A full proof can be found in Appendix~\ref{sec:proof}.
First, by construction each $G^n$ defines a distribution $X^n_{i, \cdot} \overset D = (G^n)^{-1}(Z)$ that in turn implies a conditional distribution $X^n_{\cdot, j} | X^n_{\cdot, -j} = x_{\cdot, -j}$.
We show these conditional distributions converge to the true conditional distribution of $X_{\cdot, j}$ given $X_{\cdot, -j} = x_{\cdot, -j}$.
Consequently, the probability of observing a higher test statistic under the approximated null distribution $\tilde X_{\cdot, j}^n \overset D = X_{\cdot, j}^n$, written $\alpha_j^n$, will converge to the probability under the true null distribution $\tilde X_{\cdot, j} | X_{\cdot, -j} = x_{\cdot, -j}$, i.e. $\alpha_{j}$.
Next, the Cesaro average of $K$ samples from an MCMC algorithm targeting $\tilde X_{\cdot, j}^n | X_{\cdot, -j}= x_{\cdot, -j}$, written $\hat \alpha_{j, K, n}$ will converge to $\alpha_j^n$ with probability $1$ as $K \to \infty$.
Combining these two convergences leads to the stated result.

Assuming the limiting p-values $\{\alpha_j\}$ satisfy the chosen multiple-hypothesis-testing assumptions, 
\cref{thm:asymptotic} specifies additional conditions that are sufficient for FDR control.
These conditions are not strictly fewer than those required for empirical model-X knockoff-based methods to control FDR, but they may be easier to satisfy adequately in practice.
For example, the condition that there exists a sequence $(G_n)_{n=1}^\infty$ converging to the true mapping $G$ is satisfied asymptotically by many flow architectures that are universal distribution approximators, including the Gaussianization Flows and Masked Autoregressive Flows used in our experiments \citep{huangNeuralAutoregressiveFlows2018a,mengGaussianizationFlows2020b, kobyzevNormalizingFlowsIntroduction2020}.
In practice, it is unlikely that an exact mapping $G$ will be learned, as doing so could require infinite training data, infinitely deep transformations, and exact nonconvex optimization.
Nonetheless, normalizing flows work extremely well in practice; \cref{thm:asymptotic} gives intuition for the good performance of \FlowSelect that we observe empirically.

\section{Experiments}
\begin{figure}
    \centering
    \includegraphics[width=\linewidth]{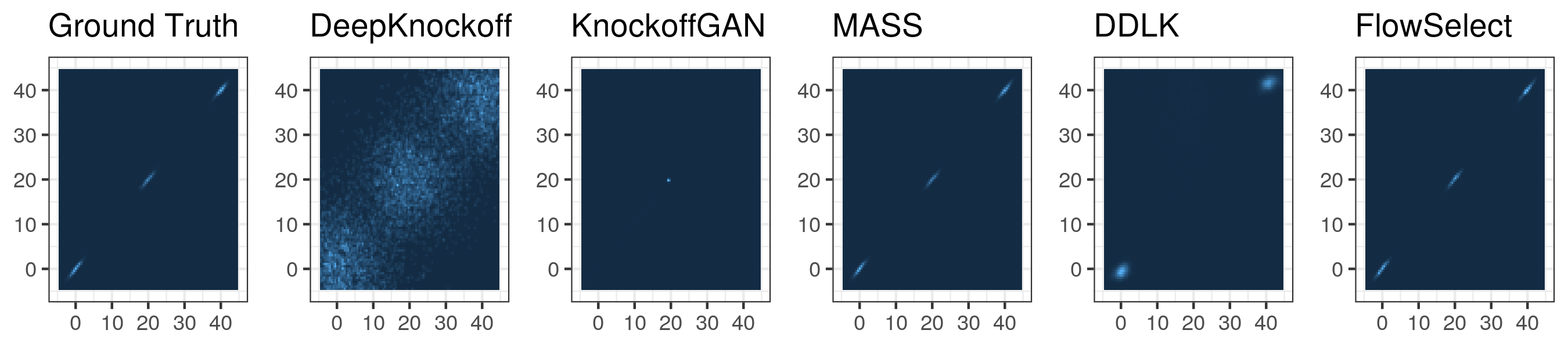}
    \caption{A density plot of the feature distribution with coordinate $j=1$ on the x-axis and coordinate $j=2$ on the y-axis. The ground truth density is compared to the normalizing flow fitted within \FlowSelect and the distribution of each knockoff method (DeepKnockoff, KnockoffGAN, MASS, and DDLK). To have FDR control, each distribution should match the distribution of the features.}
    \label{fig:linear_knockoff_distributions}
\end{figure}
\label{sec:experiments}

\subsection{Synthetic experiment with a mixture of highly correlated Gaussians}\label{sec:gaussian}

We compare \FlowSelect to the aforementioned knockoff methods with synthetic data drawn from a mixture of three highly correlated Gaussian distributions with dimension $D=100$.\footnote{Software to reproduce our experiments is available at \href{https://github.com/dereklhansen/flowselect}{https://github.com/dereklhansen/flowselect}.}
For each knockoff method, we use the exact implementation described in their respective papers, and we utilize the code made publicly available by the authors
(c.f. \cref{sec:appendix_competing_methods} for further details).
For further comparison, we also implement the MASS knockoff procedure from \citet{giminezKnockoffsMassNew2019} and the RANK knockoff procedure from \citet{fanRANKLargeScaleInference2020}.
These methods estimate the unknown feature distribution using either a mixture of Gaussians (MASS) or a sparse precision matrix (RANK), and then sample the knockoffs directly as in \citet{candesPanningGoldModelX2018}.

To generate the data, we draw $N=100,000$ highly correlated samples. For $i=1,\ldots,N$, we sample
\begin{align}
X_i &\overset {\text{i.i.d}} \sim \mathcal \sum_{m=1}^3 \pi_m p_{\mathcal N} (X_i; \mu_m, \Sigma_m),
\end{align}
with mixing weights $\pi = (0.371, 0.258, 0.371)$, mean vector $\mu = (0, 20, 40)$, and covariance matrices $\Sigma_m$.
Each covariance $\Sigma_m$ follows an AR(1) pattern such that $(\Sigma_m)_{i, j} = \rho_m^{|i-j|}$ where $\rho = (0.982, 0.976, 0.970)$.
The response $Y_i$ is linear in $f_i(X_i)$ for some function $f_i$ and coefficient vector $\beta$ i.e., $Y_i = f_i(X_i)\beta + \epsilon_i$.
Each coefficient $\beta_j$ 
equals $\frac {100} {\sqrt{N}} B_j$,
where $B_j = 0$ with probability $0.8$, $B_j = 1$ with probability 0.1, and $B_j = -1$ with probability 0.1.
We consider two different schemes for the $f_i$ that connect the features to the response.
In our linear setting, $f_i$ is equal to the identity function.
In our nonlinear setting, $f_i(x)$ is set equal to $\sin(5x)$ for odd $i$ and $f_i(x) = \cos(5x)$ for even $i$.

The experimental setting we have described so far is adapted from \citet{sudarshanDeepDirectLikelihood2020}.
However, we found that the $N=2000$ they used was too few observations for any of the methods to do well in a general non-linear setting.
Moreover, in many situations where controlled feature selection is deployed, neighboring features will be highly correlated.
To reflect this, we also increased the base correlation between features within each mixture to create a more challenging example.
We show results under the original settings of \citet{sudarshanDeepDirectLikelihood2020} in \cref{sec:ddlk_settings}.

For each model, we use 90\% of the data for training to generate null features and the remaining 10\% for calculating the feature statistics.
To define the feature statistics, we use the holdout randomization test (HRT) described at the end of \cref{sec:background}.
For the HRT, we employ different predictive models for each response type (``linear'' and ``nonlinear''). 
Specifically, for the linear response, we use the predictive log-likelihood from the LASSO \citep{tibshirani96regression}, and for the nonlinear response, we use the predictive negative mean-squared error from a random forest regressor~\citep{breiman2001random}.

First, we look at how each procedure models the covariate distribution in \cref{fig:linear_knockoff_distributions}.
In order to be valid knockoffs, the distribution of two knockoff features needs to be equal to that of the covariates.
In this challenging example, each of the empirical knockoff methods fails to match the ground truth.
In particular, DDLK and DeepKnockoffs are over-dispersed, while KnockoffGAN suffers from mode collapse. These findings for DeepKnockoffs and KnockoffGAN are similar to those reported by \citet{sudarshanDeepDirectLikelihood2020}.
Other than MASS, which directly fits a mixture of Gaussians,
\FlowSelect is the only method that matches the basic structure of the ground truth.

\begin{figure}
    \centering
    \begin{tabular}{cc}
    Mixture-of-Gaussians & scRNA-seq \\
    \includegraphics[width=0.485\linewidth]{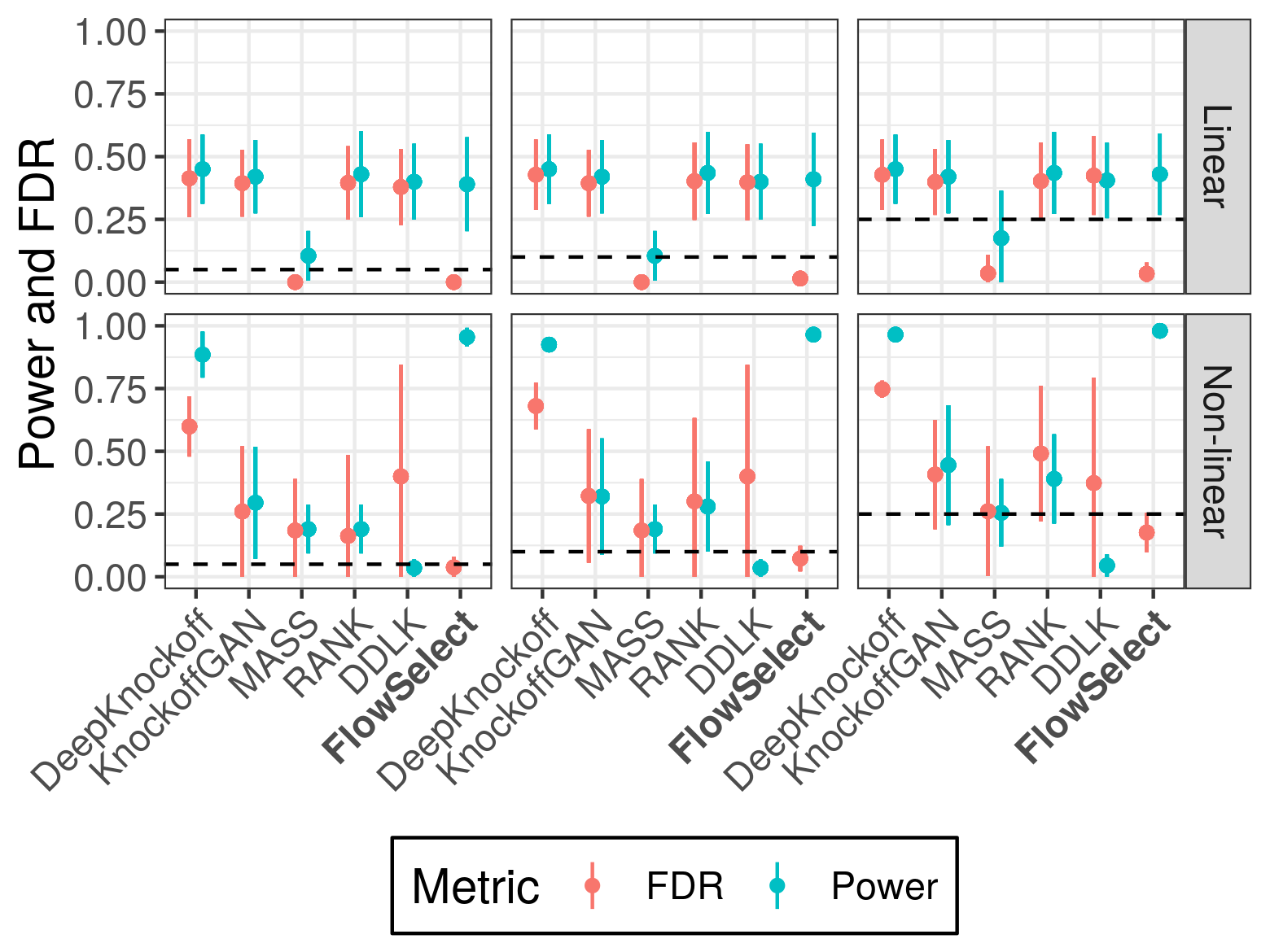}&\includegraphics[width=0.485\linewidth]{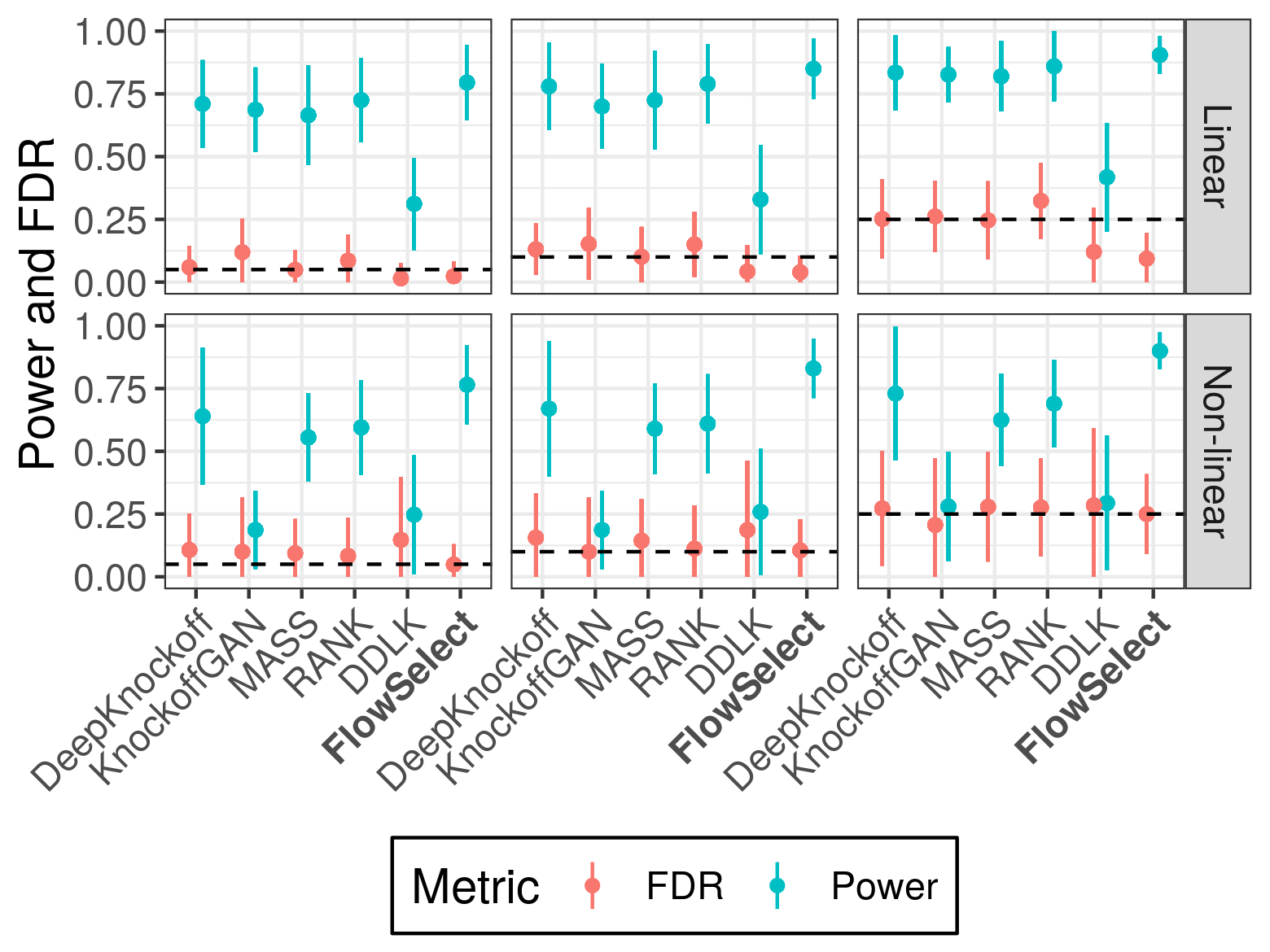}
    \end{tabular}
    \caption{Comparison of power and false discovery rate (FDR) control of \FlowSelect to knockoff methods on the Mixture-of-Gaussians dataset (left) and the scRNA-seq dataset (right) at targeted FDRs of 0.05, 0.1, and 0.25 (indicated by the dashed lines).
      Each point indicates the mean power and FDR across 20 replications and the error bars span one standard deviation either direction.
      In the top row, the response depends linearly on the features, and the feature statistics are calculated using the HRT with the LASSO.
      In the bottom row, the response depends non-linearly on the features, and the feature statistics are calculated using the HRT with random forest regression.
    }
    \label{fig:linear_fdr_boxplot}
\end{figure}
\cref{fig:linear_fdr_boxplot} shows that the empirical knockoff procedures fail to control the FDR for both linear and nonlinear responses.
One explanation for this lack of FDR control is the inability of the deep-learning-based methods to accurately model a knockoff distribution (c.f., \cref{fig:linear_knockoff_distributions}).
As a result, the assumptions for the knockoff procedure will not hold, and FDR control is not guaranteed.

The effects of misspecification are clearly visible in the case of RANK, which approximates the mixture-of-Gaussians data with a multivariate Gaussian.
However, even MASS, when given access to the correct  data distribution, does not achieve across-the-board FDR control.
This highlights the potential sensitivity of knockoffs to parameter misfit even when the underlying distributional family of the features is known.
This is confirmed by the fact that, when provided with the true parameters, the oracle Model-X maintains FDR control, though with significantly less power than \FlowSelectNS.
(c.f. Appendix \ref{sec:oracle-model-x}).




\subsection{Semi-synthetic experiment with scRNA-seq data}\label{sec:rnasq}

In this experiment, we use single-cell RNA sequencing (scRNA-seq) data from 10x Genomics \citep{rnasq_tenx}.
Each variable $X_{n, g}$ is the observed gene expression of gene $g$ in cell $n$.
These data provide an experimental setting that is both realistic and, because gene expressions are often highly correlated, challenging.
More background information about scRNA-seq data can be found in \citet{agarwal2020data}.

We normalize the gene expression measurements to lie in $[0, 1]$,
and we add a small amount of Gaussian noise so that the data is not zero-inflated.
As in the semi-synthetic experiment from \citet{sudarshanDeepDirectLikelihood2020}, we pick the 100 most correlated genes to provide a challenging, yet realistic example.
We simulate responses that are both linear and nonlinear in the features.
\cref{fig:linear_fdr_boxplot} shows that \FlowSelect maintains FDR control across multiple FDR target levels, feature statistics, and generated responses.
In cases in which the knockoff methods control FDR successfully, \FlowSelect has higher power in discovering the features the response depends on.

An advantage of knockoffs over CRT-based methods like \FlowSelect is that the predictive model only needs to be evaluated once.
Hence, while \FlowSelect has a faster runtime than DDLK for this experiment, it is slower than DeepKnockoff and KnockoffGAN.
However, \cref{fig:linear_fdr_boxplot} shows that these two models fail to reliably control FDR and have much less power than \FlowSelectNS; it is not clear how additional computational resources could be leveraged to improve the performance of these competing methods.
A full table of runtimes on the scRNA-seq dataset can be found in \cref{sec:runtimes}.

The need to compute a different predictive model for each feature within the CRT is mitigated by using efficient feature statistics such as the HRT \citep{tanseyHoldoutRandomizationTest2019} and the distilled CRT \citep{liuFastPowerfulConditional2020}.
These methods fit a larger predictive model once, then evaluate either the residuals or test mean-squared-error for each feature individually.
Moreover, the ability to scale to large feature dimensions $D$ is more limited by fitting the feature distribution than computational burden, a trait shared by both knockoff- and CRT-based methods.

\FlowSelect provides asymptotic guarantees of FDR control assuming sufficient MCMC samples have been drawn for the p-values to converge.
In this experiment, the consequence of terminating MCMC sampling before convergence is low power, rather than loss of FDR control (see \cref{fig:asymptotic_rnasq} in \cref{sec:mcmc_power_fdr}).
Even for small numbers of MCMC samples, the FDR stabilizes below the target rate, while the power steadily increases with the number of samples.
Because the MCMC run is initialized at the true features,
we speculate that the sampled features will be highly correlated with the true features in the beginning of the run, making it harder to reject the null hypothesis that a feature is unimportant.

\subsection{Ablation Study} \label{sec:ablation_study}

\textsc{FlowSelect} differs from the competing knockoff-based approaches in two ways: using normalizing flows with MCMC to model the feature distribution for sampling null features and using the CRT for feature selection.
To illustrate the impact of each of these components separately,
we compare to the procedure used in \citet{tanseyHoldoutRandomizationTest2019}, which uses mixture density networks (MDNs) to model the complete conditional distribution of each feature $\mathcal P (X_j | X_-j)$ separately.
They then sample null features from these learned distributions directly and use the HRT for feature selection.
Since both \FlowSelect and this procedure utilize the HRT,
this allows us to evaluate whether the performance improvement of \FlowSelect over empirical knockoffs is solely due to use of the HRT.

We compare the MDN-based approach to \FlowSelect on the mixture-of-Gaussians (\cref{sec:gaussian}) and scRNA-seq (\cref{sec:rnasq}) datasets.
A plot of this comparison can be found in \cref{sec:hrt_ablation}.
While the MDN-based approach was able to match the performance of \FlowSelect on the scRNA-seq dataset, it failed to control FDR at any level on the Mixture-of-Gaussians dataset, indicating that MDNs are less flexible than normalizing flows.
In aggregate, these results show that both the normalizing flows paired with MCMC and the use of the HRT for significance testing are key to the performance of \textsc{FlowSelect}.

\subsection{Real data experiment: soybean GWAS}

\label{sec:soybean}
Genome-wide association studies are a way for scientists to identify genetic variants (single-nucleotide polymorphisms, or SNPs) that are associated with a particular trait (phenotype).
We tested \FlowSelect on a dataset from the SoyNAM project \citep{song2017soynam}, which is used to conduct GWAS for soybeans.
Each feature $X_{j}$ takes on one of four discrete values, indicating whether a particular SNP is homozygous in the non-reference allele, heterozygous, homozygous in the reference allele, or missing.
A number of traits are included in the SoyNAM data; we considered oil content (percentage in the seed) as the phenotype of interest in our analysis.
There are 5,128 samples and 4,236 SNPs in total.

To estimate the joint density of the genotypes, we used a discrete flow \citep{tranDiscreteFlowsInvertible2019}.
Modeling of genomic data is typically done with a hidden Markov model \citep{Xavier2016}; however, such a model may fail to account for long range dependence between SNPs, which a normalizing flow is better suited to handle.
Having a more flexible model of the genome enables \FlowSelect to provide better FDR control for assessing genotype/phenotype relationships.
For the predictive model, we used a feed-forward neural network with three hidden layers.
Additional details of training and architecture are presented in Appendix~\ref{sec:soybean_training}.

\begin{figure}
    \centering
    \includegraphics[width=1.0\linewidth]{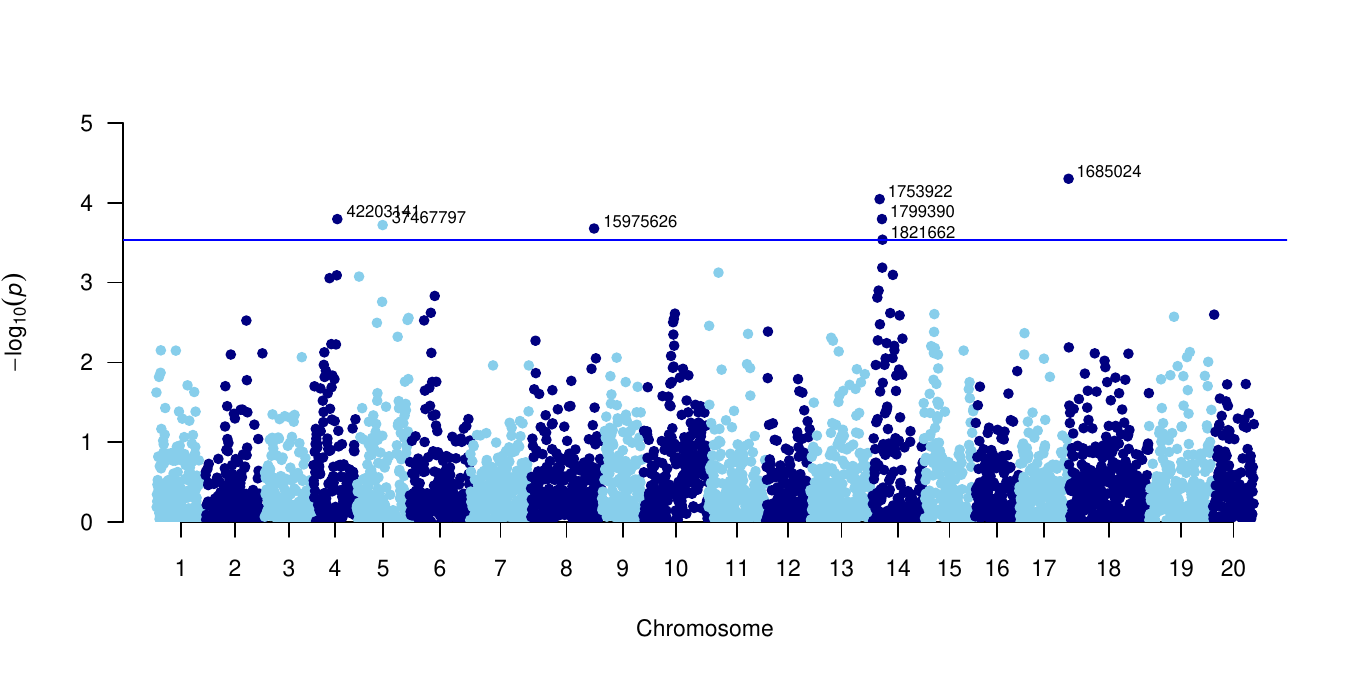}
    \caption{Manhattan plot for oil content in soybean GWAS experiment \citep{qqman}. $p$ is the estimated p-value from the \FlowSelect procedure, and the blue line indicates the rejection threshold for a nominal FDR of $20\%$.
    }
    \label{fig:manhattan-oil}
\end{figure}

A graphical representation of our results is shown as a Manhattan plot in \cref{fig:manhattan-oil}, which plots the negative logarithm of the estimated p-values for each SNP.
At a nominal FDR of 20\%, we identified seven SNPs that are associated with oil content in soybeans.
We cross-referenced our discoveries with other publications to identify SNPs that have been previously shown to be associated with oil content in soybeans.
For example, \FlowSelect identifies one SNP on the 18th chromosome, \verb+Gm18_1685024+,  which is also selected in \citet{liu_phenotype_2019}.
\FlowSelect also selects a SNP on the 5th chromosome, \verb+Gm05_37467797+, which is near two SNPs (\verb+Gm05_38473956+ and \verb+Gm05_38506373+) identified in \citet{Cao2017} but which are not in the SoyNAM dataset.
\citet{Sonah2014} identifies eight SNPs near the start of the 14th chromosome, and we select multiple SNPs in a nearby region on the 14th chromosome (seen in the peak of dots on chromosome 14 in~\cref{fig:manhattan-oil}).
However, the dataset in \citet{Sonah2014} is much larger ($\approx47,000$ SNPs), which prevents an exact comparison.
A list of all SNPs selected by our method is provided in Appendix~\ref{sec:soybean_training}.
For this experiment, \FlowSelect tests over $4000$ features in 10 hours using a single GPU.
None of the empirical knockoff procedures \citep{sudarshanDeepDirectLikelihood2020, jordon2018knockoffgan, romanoDeepKnockoffs2018} tested more than $387$ features.
This shows the potential for \FlowSelect for high-dimensional feature selection with FDR control in a reasonable amount of time.
Additional details about this experiment are available in Appendix~\ref{sec:soybean_training}.

\section{Discussion}

\FlowSelect enables scientists and other practitioners to discover features a response depends on while controlling false discovery rate, using an arbitrary predictive model; even large-scale nonlinear machine learning models can be utilized.
By making fewer false discoveries for a fixed sensitivity level, \FlowSelect can reduce the cost of follow-up experiments by limiting the number of irrelevant features considered.
In contrast to the original model-X knockoffs method, \FlowSelect does not require the feature distribution to be known a priori, nor does it require the feature distribution to have a particular form (e.g., Gaussian). Neither of these conditions are often satisfied in practice.

One limitation shared by both the conditional randomization test (CRT) and knockoffs is low power in cases in which important features are highly correlated with other important features.
To mitigate this limitation, the CRT can be applied to test the significance of groups of correlated features rather than individual features.
Within the \FlowSelect framework, this entails modifying the MCMC step to draw null samples of groups of features conditioned on the others.
The group's p-value can then be calculated with the same holdout randomization test (HRT) statistic used for testing individual features.
Group feature selection has also been explored for knockoffs \citep{daiKnockoffFilterFDR2016,liuFastPowerfulConditional2020}.

Another limitation of \FlowSelect stems from its reliance on normalizing flows.
The flexibility of normalizing flows, though often beneficial, comes at a cost:
sufficient training examples are needed to learn the feature distribution, limiting applicability in data-starved regimes.
Fortunately, as we show in \cref{sec:ddlk_settings}, \FlowSelect fares no worse than competing methods in low-data settings.
In these regimes, \FlowSelect could also use other density estimation techniques such as autoregressive models.

Furthermore, learning the feature distribution (potentially from limited data) is not the sole difficulty that the deep-learning-based knockoff methods face.
To demonstrate that there are additional sources of difficult for knockoff-based methods, we gave DDLK, which typically fits the data distribution as part of its training procedure, access to the \textit{the exact joint density}; neither the empirical FDR nor the power improved significantly (c.f. \cref{sec:ddlk_oracle}).
This result points to a failure of DDLK to enforce the swap property, which is a challenging task as the number of swaps grows exponentially with the number of features.
\FlowSelect, on the other hand, achieves FDR control under a different set of conditions that often are simpler to satisfy adequately in practice.

\begin{ack}
Derek Hansen acknowledges support from the National Science Foundation Graduate Research Fellowship Program under grant no. 1256260. 
Any opinions,
findings, and conclusions or recommendations expressed in this material are those of the
author(s) and do not necessarily reflect the views of the National Science Foundation.
\end{ack}

\bibliography{mcmc_crt}
\bibliographystyle{icml2022}

\section*{Checklist}

\begin{enumerate}

\item For all authors...
\begin{enumerate}
  \item Do the main claims made in the abstract and introduction accurately reflect the paper's contributions and scope?
    \answerYes{}
  \item Did you describe the limitations of your work?
    \answerYes{Runtime discussion in section 5.2 and limitations of normalizing flows in section 6}
  \item Did you discuss any potential negative societal impacts of your work?
    \answerNA{}
  \item Have you read the ethics review guidelines and ensured that your paper conforms to them?
    \answerYes{}
\end{enumerate}

\item If you are including theoretical results...
\begin{enumerate}
  \item Did you state the full set of assumptions of all theoretical results?
    \answerYes{Theorem 1 statement lists all assumptions.}
        \item Did you include complete proofs of all theoretical results?
    \answerYes{Yes; Appendix B contains the proof to Theorem 1.}
\end{enumerate}

\item If you ran experiments...
\begin{enumerate}
  \item Did you include the code, data, and instructions needed to reproduce the main experimental results (either in the supplemental material or as a URL)?
    \answerYes{Code included as supplement, and specific details were described in the Appendix.}
  \item Did you specify all the training details (e.g., data splits, hyperparameters, how they were chosen)?
    \answerYes{In Appendices D and E.}
        \item Did you report error bars (e.g., with respect to the random seed after running experiments multiple times)?
    \answerYes{All figures comparing FDR and Power results (Figure 2 in main text, Figures 4-6, 8 in the Appendix) contain error bars indicating results across runs.}
        \item Did you include the total amount of compute and the type of resources used (e.g., type of GPUs, internal cluster, or cloud provider)?
    \answerYes{We specify hardware used in Appendix F.}
\end{enumerate}

\item If you are using existing assets (e.g., code, data, models) or curating/releasing new assets...
\begin{enumerate}
  \item If your work uses existing assets, did you cite the creators?
    \answerYes{See Appendix C}
  \item Did you mention the license of the assets?
    \answerYes{See Appendix C}
  \item Did you include any new assets either in the supplemental material or as a URL?
    \answerNA{}
  \item Did you discuss whether and how consent was obtained from people whose data you're using/curating?
    \answerNA{Data are publicly available}
  \item Did you discuss whether the data you are using/curating contains personally identifiable information or offensive content?
    \answerNA{}
\end{enumerate}

\item If you used crowdsourcing or conducted research with human subjects...
\begin{enumerate}
  \item Did you include the full text of instructions given to participants and screenshots, if applicable?
    \answerNA{}
  \item Did you describe any potential participant risks, with links to Institutional Review Board (IRB) approvals, if applicable?
    \answerNA{}
  \item Did you include the estimated hourly wage paid to participants and the total amount spent on participant compensation?
    \answerNA{}
\end{enumerate}

\end{enumerate}


\clearpage
\appendix

\section{Normalizing flows} \label{sec:flows}
Normalizing flows \citep{papamakariosNormalizingFlowsProbabilistic2019a} represent a general framework for density estimation of a multi-dimensional distribution with arbitrary dependencies.
Briefly, suppose $X \sim \mathcal P_{X}$ is a random variable in $\mathbb R^{d}$.
Now, let $Z \sim \mathcal N(0, I_{d})$ be a multivariate standard normal distribution.
We assume there exists a mapping $G$ that is triangular, increasing, and differentiable such that
\[
  G(X) = Z.
\]
A formal treatment of when such a $G$ exists can be found in \citet{bogachevTriangularTransformationsMeasures2005}.
However, a sufficient condition is that the density of $X$ is greater than $0$ on $\mathbb R^{d}$ and the cumulative density function of $X_{j}$, conditional on the previous components $X_{\le j}$,  is differentiable with respect to $X_{j}, X_{\le j}$ \citep{papamakariosNormalizingFlowsProbabilistic2019a}:
\begin{equation*}
  U_{i} = G_{i}(X) \equiv F_{i}(X_{i}| X_{\le i})
\end{equation*}
From this construction, each $U_{i}$ is independent of all previous $U_{i}$ and has distribution $\mathrm{Unif}[0, 1]$.
From there, we simply set $Z_{i} = \Phi^{-1}(U_{i})$, where $\Phi$ is the CDF of the standard normal.

Since $G_{i}(X)$ depends only on the elements in $X$ up to $i$, it is triangular.
Because $p_{X} > 0$, the conditional cdfs are strictly increasing, so $G$ is an increasing map.
Finally, since each cdf is differentiable, the entire map $G$ is differentiable, and its Jacobian is non-zero.

Because of the inverse mapping theorem, $G$ is invertible and we can write
\[
  X = G(Z).
\]

Normalizing flows are a collection of distributions that parameterize a family of invertible, differentiable transformations $G_{\theta}$ from a fixed base distribution $Z$ to an unknown distribution $X$.
Using the change-of-variables theorem, we can express the distribution of $X$ in terms of the base distribution density $p_{Z}$ and the transformation $G_{\theta}$:
\[
 p_{\theta}(X) = p(G_{\theta}(X)) \left | \det \left( \frac{\partial G_{\theta}(X)}{\partial X} \right ) \right |
\]
where $\frac{\partial G_{\theta}(X)}{\partial X}$ is the Jacobian of $G$.
The goal is to find a parameter value $\hat \theta$ that maximizes the likelihood of the observed $X$:
\[
 \hat \theta = \argmax_{\theta} p_{\theta}(X).
\]

A key feature of normalizing flows is that they are composable.

\subsection{Flow Architecture}
In experiments, the first layer $G$ is a Gaussianization flow \citep{mengGaussianizationFlows2020b} applied elementwise:
\[
  G_{j} (X_{j}) = \Phi^{-1} \left (\sum_{m=1}^{M}\sigma \left( \frac{X_{j} - \mu_{j,m}}{s_{j, m}} \right) \right),
\]
where $\Phi^{-1}$ is the standard normal inverse CDF.
With sufficiently large $M$, this Gaussianization layer can approximate any univariate distribution.
This is composed with a Masked Autoregressive Flow (MAF) $F$ \citep{papamakariosMaskedAutoregressiveFlow2017}, which consists of MADE layers interspersed with batch normalization and reverse permutation layers:
\[
  \begin{split}
    \textrm{MADE}_{j, k} &= (X_{j} - \mu_{j,k}) \exp (-\alpha_{j,k}) \\
    \text{where}~\mu_{j} &= f_{\mu_{j,k}}(X_{< j}) \\
    \alpha_{j} &= f_{\alpha_{j, k}}(X_{< j}) \\
    F &= \textrm{MADE}_{j, K} \circ \textrm{BatchNorm} \circ \textrm{Reverse} \circ \textrm{MADE}_{j, K-1} \circ \cdots \textrm{BatchNorm} \circ \textrm{Reverse} \circ \textrm{MADE}_{j, 1}
\end{split}
\]
Here, $f_{\mu_{j}}$ and $f_{\alpha_{j}}$ are fully connected neural networks.

\section{Proof of convergence}\label{sec:proof}
\asymptotic*
\begin{proof}[Proof of Theorem 1]
Without loss of generality, we consider the first feature, which is indexed by $j = 1$.
Let $p_X$ be the density of each row of the matrix $X_{i, \cdot}$ and $p_Z$ the density of the base variable $Z$.
For each i.i.d observation at $i=1, \dots, N$, we define $F$ to be the cumulative distribution function of $X_{i, 1}$ conditional on the other features $X_{i, -1} = x_{i, -1}$:
\begin{equation}
  \begin{split}
  F(x_{1}) \triangleq \mathcal P(X_{i, 1} \le x_{1} \vert X_{i,-1} = x_{i, -1}) &= \frac{\int_{-\infty}^{x_{1}} p_{X}(x^{\prime}_{1}, x_{i, -1}) d x^{\prime}_{1}} {\int_{-\infty}^{\infty} p_{X}(x^{\prime}_{1}, x_{i, -1}) d x^{\prime}_{1}} \\
  &= \frac{\int_{-\infty}^{x_{1}} p_{Z}(G(x^{\prime}_{1}, x_{i, -1})) |\partial G(x^{\prime}_{1}, x_{i, -1})| d x^{\prime}_{1}} {\int_{-\infty}^{\infty} p_{Z}(G(x^{\prime}_{1}, x_{i, -1})) |\partial G(x^{\prime}_{1}, x_{i, -1})| d x^{\prime}_{1}}. \\
\end{split}
\end{equation}
For a particular mapping $G^n$, we define $F^{n}$ analogously:
\begin{equation}
  \begin{split}
  F^{n} (x_{1})&\triangleq \frac{\int_{-\infty}^{x_{1}} p_{Z}(G^{n}(x^{\prime}_{1}, x_{i, -1})) |\partial G^{{n}}(x^{\prime}_{1}, x_{i, -1})| d x^{\prime}_{1}} {\int_{-\infty}^{\infty} p_{Z}(G^{{n}}(x^{\prime}_{1}, x_{i, -1})) |\partial G^{{n}}(x^{\prime}_{1}, x_{i, -1})| d x^{\prime}_{1}}.
\end{split}
\end{equation}
Since $G^{{n}}$ and $G$ are continuously differentiable,
\begin{equation}
p_{Z}(G^{n}(x^{\prime}_{1}, x_{i, -1})) |\partial G^{{n}}(x^{\prime}_{1}, x_{i, -1})| \to p_{Z}(G(x^{\prime}_{1}, x_{i, -1})) |\partial G(x^{\prime}_{1}, x_{i, -1})|~\text{as}~n\to\infty.
\end{equation}
Then, by the dominated convergence theorem, $F^{n} \to F$ pointwise.

Let $X_{i, 1}^n \sim F^n$.
Since $F^{n} \to F$ pointwise, and $F$ is a distribution function,
\(X_{i, 1}^{n}\)
converges in distribution to
\(X_{i, 1} \mid X_{i,-1} = x_{i, -1}\).
Likewise, the joint distribution across all independent observations, written $X_{\cdot, 1}^n$, converges in distribution to $X_{\cdot, 1} \mid X_{\cdot, -1} = x_{\cdot, -1}$.

Now, let $\tilde X^n_{\cdot, 1}$ be equal in distribution to $ X^n_{\cdot, 1}$, but sampled such that it is independent of the outcome $Y$.
It follows from the reasoning above that $\tilde X^n_{\cdot, 1}$ converges to the desired null distribution $\tilde X_{\cdot, 1} \vert X_{\cdot, -1}$ as $n \to \infty$.
Define $g_{1} (\tilde x_{\cdot, 1}) \triangleq 1[T_{1}(X) < T_{1}([\tilde x_{\cdot, 1}, X_{\cdot, -1}])]$.
With the regularity condition that $T_1$ is discontinuous on a set of measure zero,
the expectation converges:
\begin{equation} \label{eq:conv_expectation}
 \lim_{n \to \infty}\mathbb E_{\tilde X_{\cdot, 1}^{n}} (g_{1}) \to \mathbb E_{\tilde X_{\cdot, 1} \vert X_{\cdot, -1} = x_{\cdot, -1}} (g_{1}) = \alpha_{1}.
\end{equation}

The Cesaro average of $g$ calculated over MCMC samples that target the distribution of $\tilde X_{\cdot, 1}^n$ under the probability law of $G_n$ converges almost surely to $\mathbb E_{\tilde X_{\cdot, 1}^{n}}(g_{1})$
\citep{smithBayesianComputationGibbs1993}. That is,
\begin{equation} \label{eq:conv_mcmc}
   \lim_{K \to \infty} \hat \alpha_{j, K, n}= \lim_{K \to \infty} \frac {1}{K} \sum_{k=1}^{K} g_{1}(\tilde X_{\cdot, 1, k}) = \mathbb E_{\tilde X_{\cdot, 1}^{n} }(g_{1}) ~w.p.1.
 \end{equation}

Combining \cref{eq:conv_expectation} and \cref{eq:conv_mcmc} gives the desired result.
\end{proof}
 \newpage

\section{Feature datasets}

\begin{tabular}{rllllll}
  Name & Covariate & Response & $N$ & $D$ & \# Relevant & Source \\
  \midrule
  Gaussian Mixture & Synthetic & Synthetic & $100,000$ & $100$ & $20$ & -\\
  scRNA-seq & Real & Synthetic & $100,000$ & $100$ & $10$ &  \citet{rnasq_tenx}\\
  Soybean & Real & Real & $5,128$ & $4,236$ & - & \citet{soynam} \\
\end{tabular}

\paragraph{Licensing} All of the data used is available for personal use.
Terms for the scRNA-seq data can be found here: \url{https://www.10xgenomics.com/terms-of-use}.
The scRNA-seq data was accessed using scvi-tools \citep{Gayoso2021scvitools}, distributed under the
BSD 3-Clause license.
The soybean data is part of the SoyNAM R package \citep{soynam}, distributed under the GPL-3 license.

\section{Architecture and training details for synthetic experiments}

\subsection{FlowSelect}

For \FlowSelect, the joint distribution was fitted with a GaussMAF normalizing flow as described in Appendix~\ref{sec:flows}.
The first Gaussianization layer consisted of $M=6$ clusters, followed by 5 layers of MAF.
Within each MAF layer, the neural network consisted of three masked fully connected residual layers with $100$ hidden units, followed by a BatchNorm layer.

We trained the Gaussianization layer first with $100$ epochs and learning rate $1 \times 10^{-3}$ within the ADAM optimizer. This allowed the Gaussianization layer to learn the marginal distribution of each feature.
Then, we jointly trained the whole architecture with $100$ epochs and learning rate $1 \times 10^{-3}$ using ADAM.

\paragraph{MCMC} We draw $1000$ samples using a Metropolis-Hastings procedure.
The proposal distribution is a random walk:
\begin{equation*}
  \label{eq:mcmc_random_walk}
  X_{i,j,k}^{*} \sim \mathcal N(\tilde X_{i,j,k-1}, \hat \sigma_{j}^{2}),
\end{equation*}
where $\hat \sigma_{j}^{2}$ is the sample conditional variance:
\begin{equation*}
  \begin{split}
    \hat \sigma_{j}^{2} &= \hat \Sigma_{j, j} - \hat \Sigma_{j, -j} \hat \Sigma_{-j, -j}^{-1} \hat \Sigma_{j, -j}^{T} \\
    \text{where}~\hat \Sigma_{j} &= \widehat{\text{Var}}(X)
\end{split}
\end{equation*}

\subsection{Variable selection methods}

\paragraph{Linear}
For the linear response, we estimate a linear model with an L1 penalty (aka the LASSO) on training data:
\begin{equation}
  \hat \beta = \argmin_{\beta} \frac{1}{N} \|X\beta - Y\|_{2}^{2} + \lambda \sum_{j=1}^{D} |\beta_{j}|
\end{equation}
The penalization term $\lambda$ is selected via 5-fold cross-validation.

\paragraph{Nonlinear} For the nonlinear response, we fit a random forest on the training data.
The hyperparameters are the defaults in the scikit-learn implementation.

\paragraph{Feature statistic}
If $\hat f(X)$ is the fitted regression function, then the feature statistic is the negative mean-squared error:
\begin{equation*}
  T(X, Y) = - \frac 1 N \|\hat f(X) - Y \|_{2}^{2} .
\end{equation*}

\subsection{Competing methods}
\label{sec:appendix_competing_methods}
For DDLK \citep{sudarshanDeepDirectLikelihood2020}, KnockoffGAN \citep{jordon2018knockoffgan}, and DeepKnockoffs, \citep{romanoDeepKnockoffs2018}, we used the exact architecture and hyperparameter settings from their respective papers.
For the ablation study in \cref{sec:ablation_study}, we use the exact implementation in \citet{tanseyHoldoutRandomizationTest2019}.
For these methods, we used the code that the researchers graciously made publicly available:

\begin{center}
\begin{tabular}{r|l}
  Method & Link \\
  \midrule
DDLK & \url{https://github.com/rajesh-lab/ddlk/} \\
  DeepKnockoffs & \url{https://github.com/msesia/deepknockoffs/} \\
  HRT (MDN) & \url{https://github.com/tansey/hrt/} \\
  KnockoffGAN & \url{https://github.com/firmai/tsgan/tree/master/alg/knockoffgan} \\
\end{tabular}
\end{center}

For MASS \citep{giminezKnockoffsMassNew2019}, we followed their described procedure and fit a mixture of Gaussians to the feature distribution using scikit-learn, selecting the number of components via the Akiake Information Criterion (AIC).
We then used the \texttt{knockoffs} R package, available on CRAN, to sample knockoffs using the estimated parameters for each component.

For RANK \citep{fanRANKLargeScaleInference2020}, we estimate the sparse precision matrix using the Graphical LASSO \citep{friedmanSparseInverseCovariance2008} implemented in sci-kit learn, using cross-validation to tune the regularization parameter. We then use the \texttt{knockoffs} R package to sample the knockoffs with this covariance.

\section{Architecture and training details for soybean GWAS} \label{sec:soybean_training}

\paragraph{Discrete flows}

For the discrete flows in the soybean example, we use a single layer of MADE which outputs a dimension of size $4$. $\mu$ is then set equal to the argmax of this output.

For training the flows, we use a relaxation of argmax with temperature equal to $0.1$.

\paragraph{Discrete MCMC}

Each feature has $K=4$ values, so we can enumerate all four possible states for each proposal and sample in proportional to these probabilities via a Gibbs Sampling procedure.
Setting the probabilities leads to an acceptance rate of $1$, and the samples are uncorrelated since the previous sample doesn't enter into the proposal distribution

\paragraph{Predictive model}
For the predictive model of each trait conditional on the SNPs, we use a fully connected neural network.
This network has three hidden layers of size 128, 256, and 128.
ReLU activations are used between each fully connected layer.
Dropout is used on both the input layer and after each hidden layer with $p=0.2$.
The learning rate in ADAM was set to $1 \times 10^{-5}$, with early stopping implemented using a held-out validation set.

The feature statistic for each sample is the negative mean-squared error (MSE) for each observation.

\paragraph{Runtime}
To obtain sufficient resolution on roughly $4200$ simultaneous tests, we drew 100,000 samples from our model. The runtime was 10 hours using a single NVIDIA 2080 Ti.

\paragraph{Selected SNPs}
\cref{tab:soybean_snps} shows the SNPs selected by \FlowSelect that are associated with oil content in soybeans.
\begin{table}[ht]
\centering
\begin{tabular}{llc}
  \toprule
Chromosome & SNP & p-value \\
  \hline
 4 & Gm04\_42203141 & 1.60e-04 \\
   5 & Gm05\_37467797 & 1.90e-04 \\
   8 & Gm08\_15975626 & 2.10e-04 \\
  14 & Gm14\_1753922 & 9.00e-05 \\
  14 & Gm14\_1799390 & 1.60e-04 \\
  14 & Gm14\_1821662 & 2.90e-04 \\
  18 & Gm18\_1685024 & 5.00e-05 \\
   \bottomrule
\end{tabular}
\vspace{1.5mm}
\caption{Selected SNPs for soybean GWAS experiment.}
\label{tab:soybean_snps}
\end{table}
\clearpage

\section{Runtime comparison of each controlled feature selection method} \label{sec:runtimes}
\begin{table}[h!]
\centering
\begin{tabular}{lr}
  \toprule
 Method & Runtime (min) \\
  \midrule
DeepKnockoff & 3.0  \\
KnockoffGAN & 3.73 \\
MASS & 12.6 \\
DDLK & 91.9  \\
\FlowSelect & 59.4 \\
   \bottomrule
\end{tabular}
\caption{The median runtime for each method on the scRNA-seq data with $D=100$ features and $N=100,000$ observations.
All experiments were implemented using PyTorch, except for KnockoffGAN, which was implemented in Tensorflow, and MASS, which we implemented using scikit-learn and the knockoffs R package.
The experiments were conducted using an Intel Xeon Gold 6130 CPU and an NVIDIA GeForce RTX 2080 Ti GPU.
}
\label{tbl:rnasq_runtimes}
\end{table}

\section{Comparison to Holdout Randomization Test of \citet{tanseyHoldoutRandomizationTest2019}} \label{sec:hrt_ablation}
\begin{figure}[!h]
  \centering
    \begin{tabular}{cc}
    Mixture-of-Gaussians & scRNA-seq \\
    \includegraphics[width=0.30\linewidth]{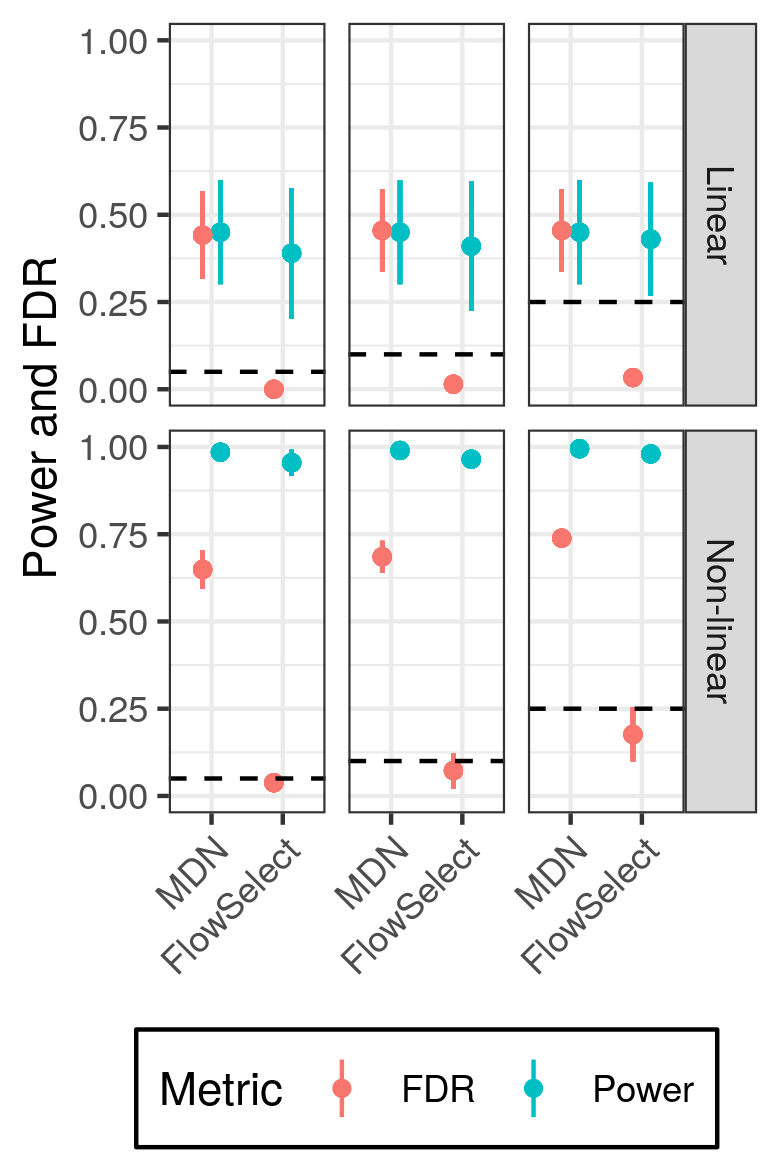}
        &\includegraphics[width=0.30\linewidth]{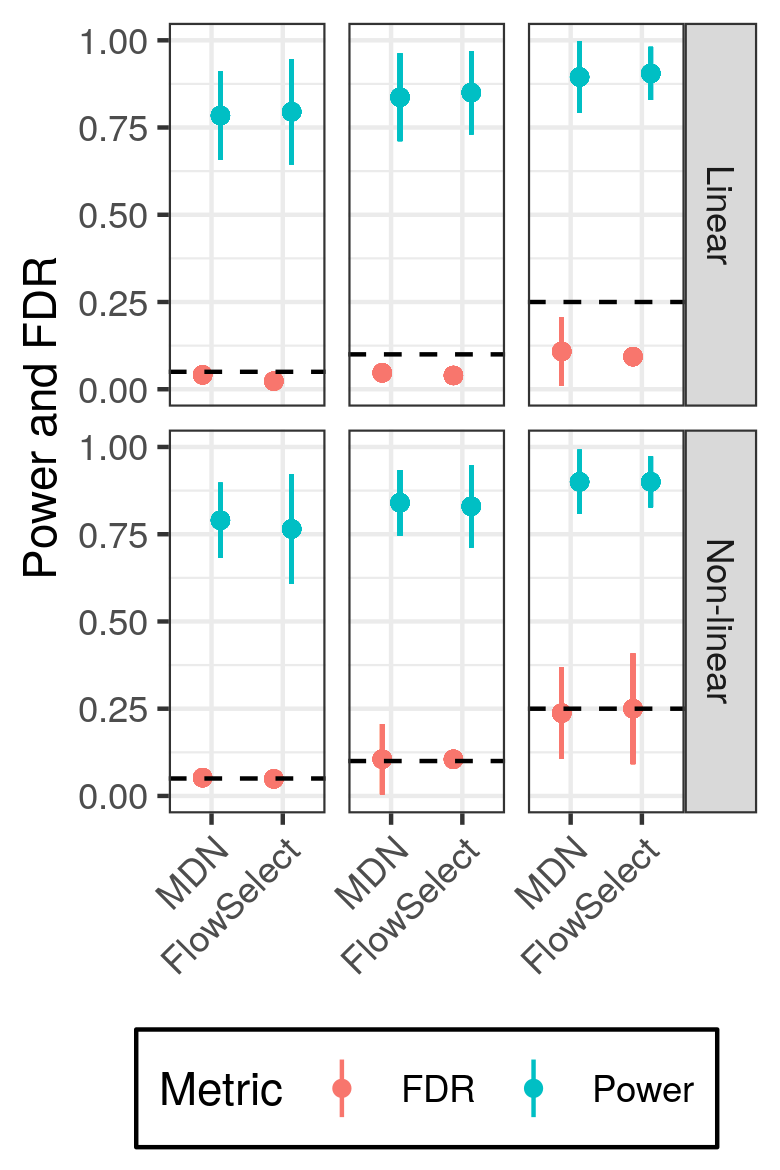}
            \end{tabular}
            \caption{Comparison of \FlowSelect to the HRT procedure in \citet{tanseyHoldoutRandomizationTest2019} which samples the complete conditionals using multiple mixture-density-networks (MDNs). Each column shows the power and observed false discovery rate (FDR) at targeted FDRs of 0.05, 0.1, and 0.25 (indicated by the dashed lines).
              The experimental settings for each dataset are the same as in \cref{fig:linear_fdr_boxplot}.}
    \label{fig:gaussian_hrt}
  \end{figure}

  \clearpage

\section{Oracle Model-X}\label{sec:oracle-model-x}

\begin{figure}[!h]
    \centering
    \includegraphics[width=0.5\textwidth]{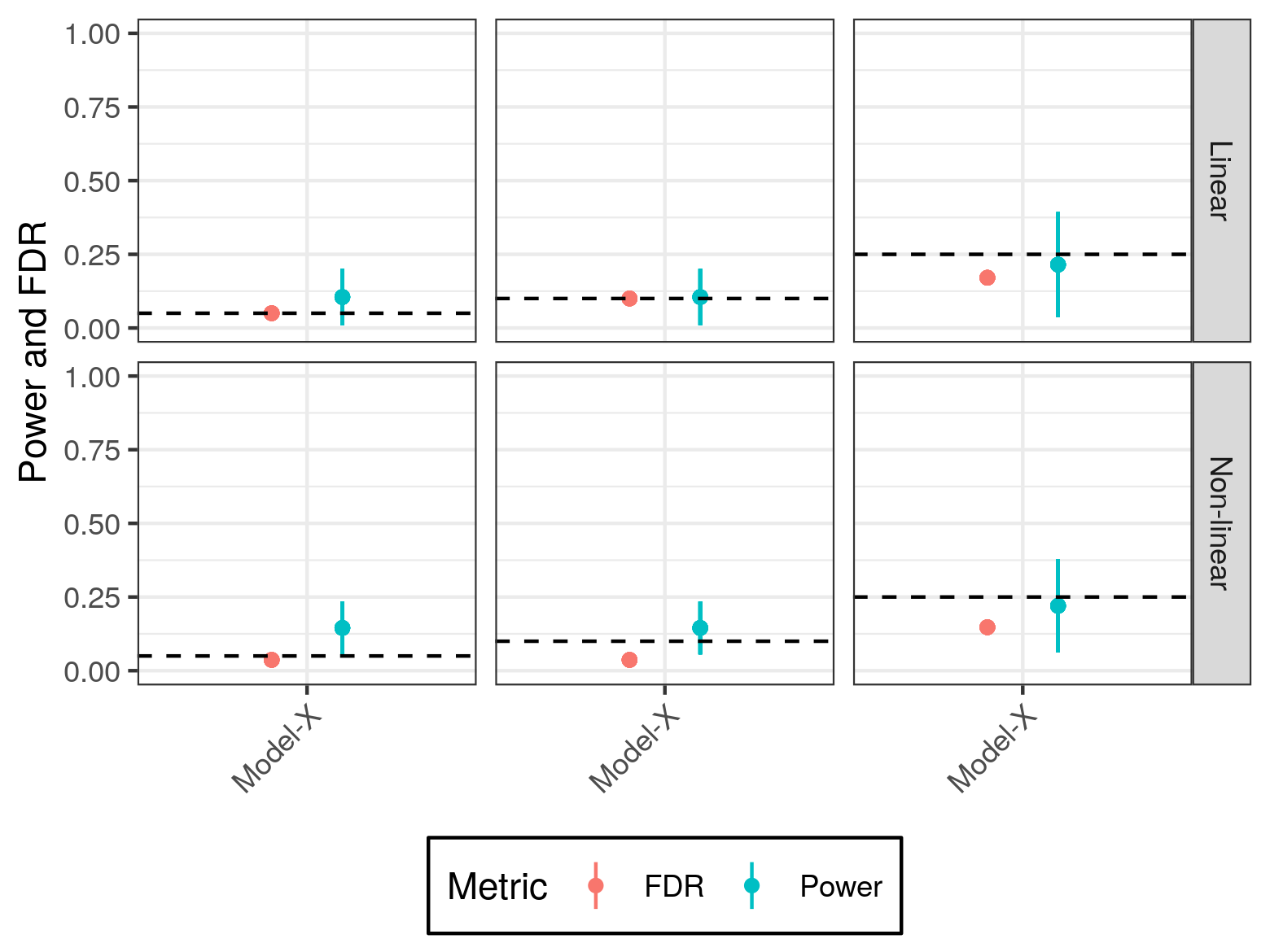}
    \caption{FDR control and power of Oracle Model-X knockoffs on the mixture-of-Gaussians dataset (compare to \cref{fig:linear_fdr_boxplot}).
    }
    \label{fig:oracle_modelx_boxplots}
\end{figure}

\section{DDLK with true joint distribution} \label{sec:ddlk_oracle}

\begin{figure}[ht]
    \centering
    \includegraphics[width=0.5\linewidth]{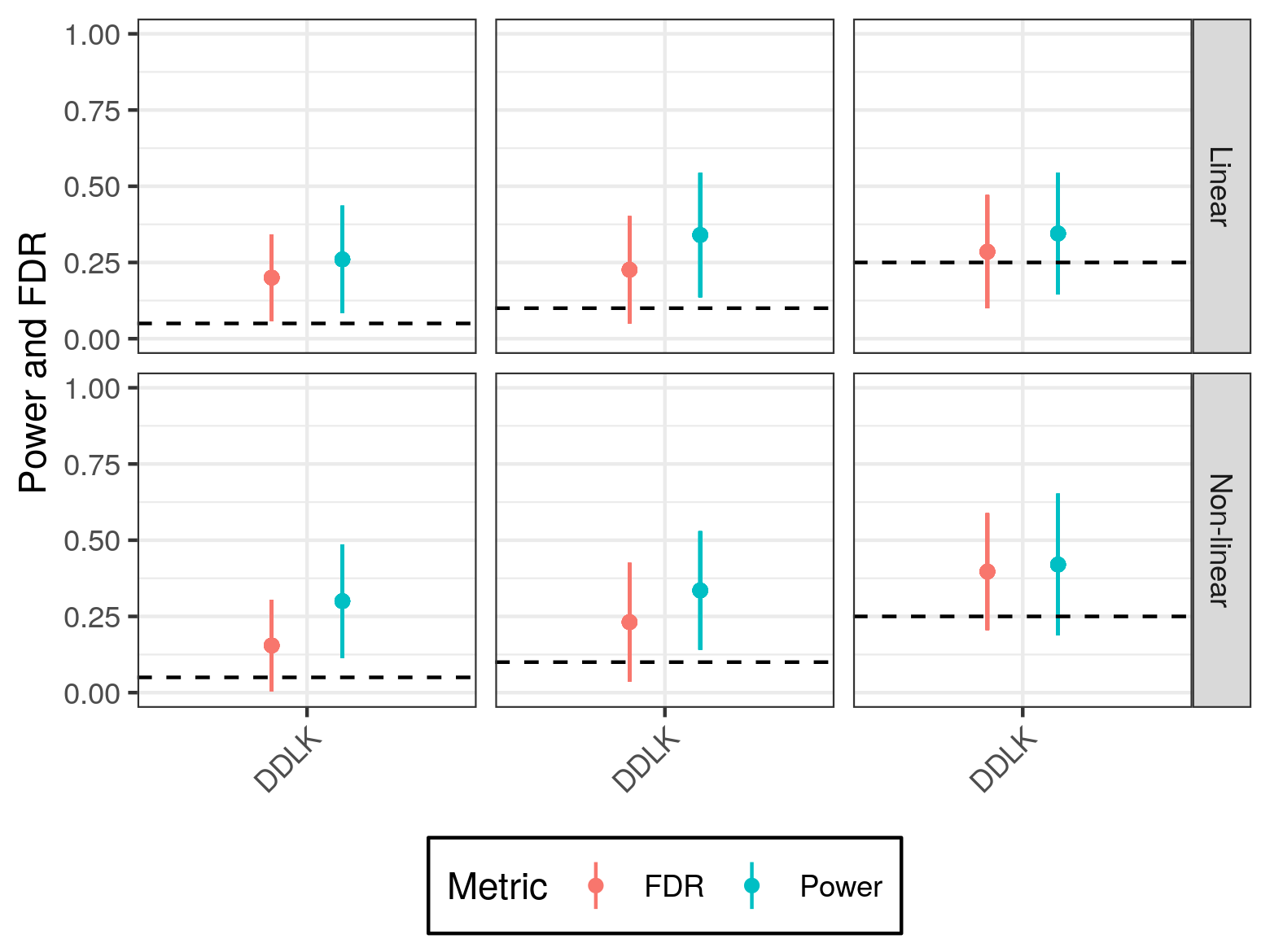}
    \caption{FDR control and power of DDLK on the mixture-of-Gaussians dataset using the ground truth feature density in training (compare to \cref{fig:linear_fdr_boxplot}).
    }
    \label{fig:ddlk_truejoint}
  \end{figure}

  \clearpage

\section{Observed Power and FDR control for given number of MCMC samples}\label{sec:mcmc_power_fdr}
\begin{figure}[!h]
    \centering
    \includegraphics[width=0.5\linewidth]{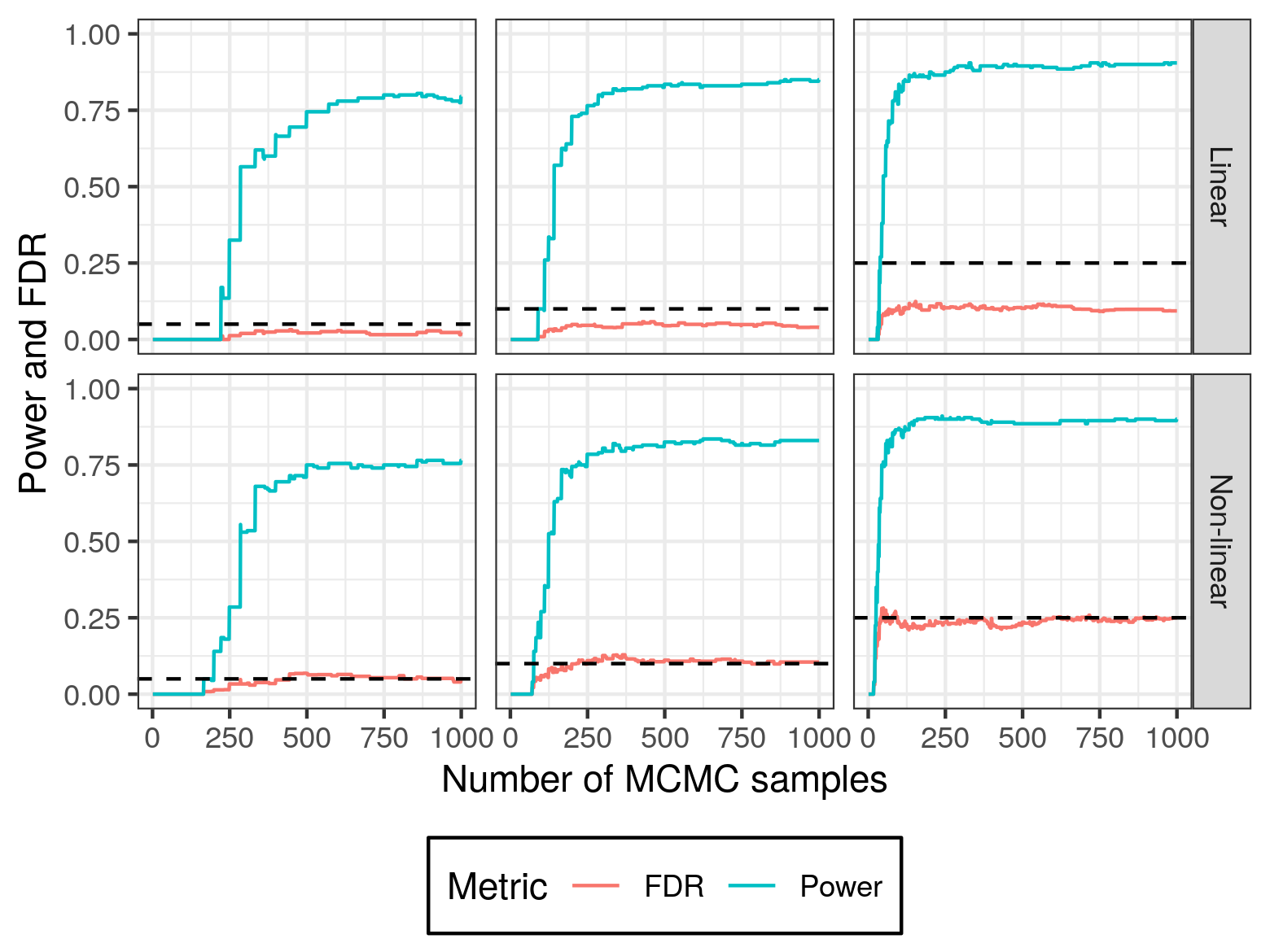}
    \caption{Power and FDR control of \FlowSelect on the scRNA-seq dataset as a function of the number of MCMC samples at targeted FDRs of 0.05, 0.1, and 0.25 (indicated by the dashed lines).
    This suggests that the consequence of terminating the MCMC chain prematurely leads to a drop in power but FDR control is still maintained.
    }
    \label{fig:asymptotic_rnasq}
\end{figure}

\section{Mixture-of-Gaussians results for FDR and Power under \citet{sudarshanDeepDirectLikelihood2020} settings}\label{sec:ddlk_settings}

\begin{figure}[!h]
\begin{center}
        \includegraphics[width=0.5\linewidth]{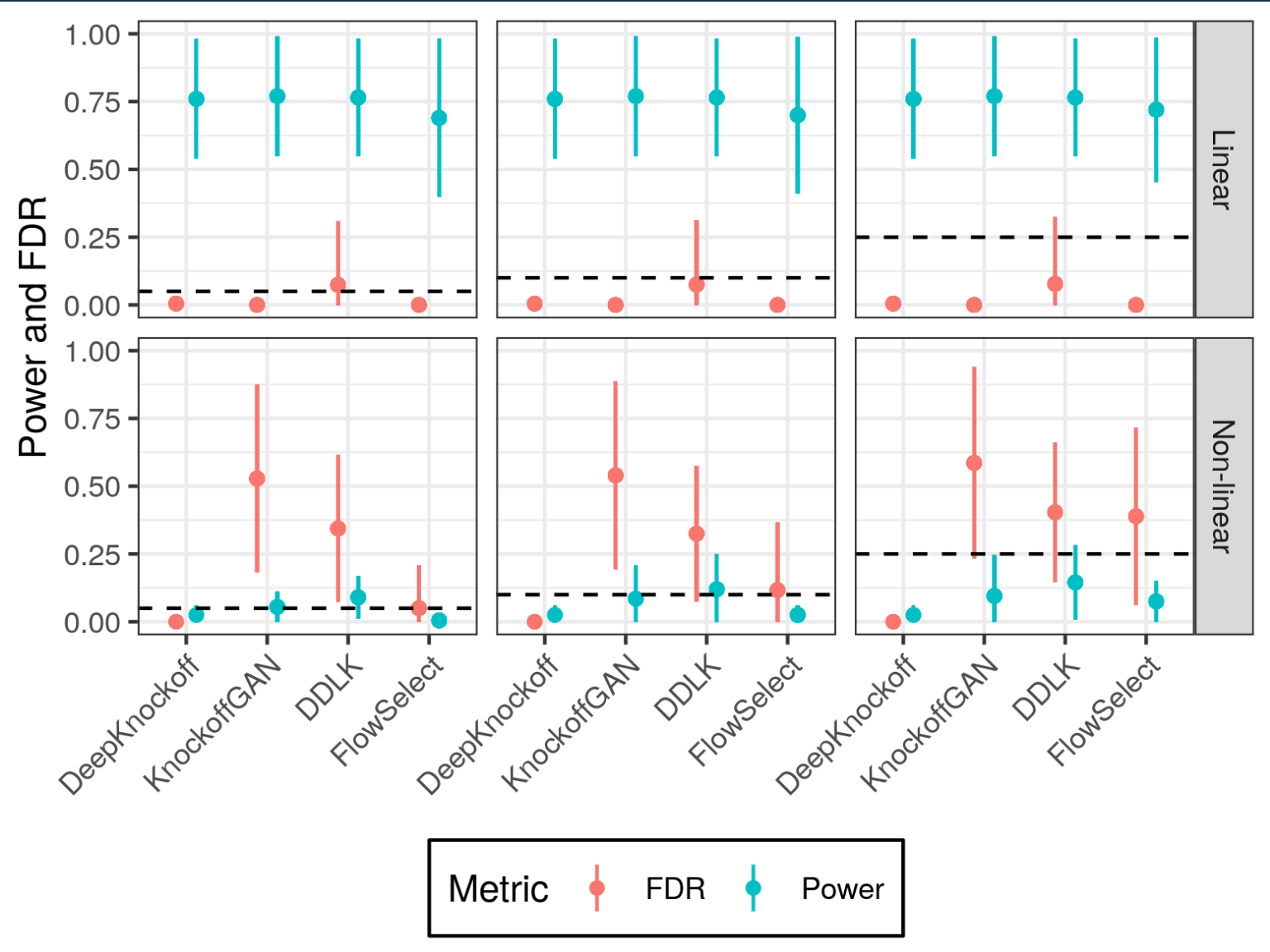}
\end{center}
    \caption{Mixture-of-gaussians setup with \(\rho = (0.6, 0.4, 0.2)\) and \(N=2000\) to match the settings in \citet{sudarshanDeepDirectLikelihood2020}.
In the linear response setting, which matches the data-generating process of \citet{sudarshanDeepDirectLikelihood2020}, all competing knockoff-based methods (i.e., DDLK, KnockoffGAN, and DeepKnockoff) as well as FlowSelect control the FDR at 5\%, 10\% and 25\% levels and achieve a power of about \(0.75\).
In the non-linear response setting, none of the methods control FDR, except for DeepKnockoffs which had nearly zero power.
The good performance in the linear setting can be explained by the LASSO feature statistic shrinking most null features to zero since they have relatively low correlation.
Since FDR control should hold for any response setting, these findings suggest that none of the methods do well in modeling the underlying distribution with $N=2000$ observations.
}
\end{figure}

\newpage
\section{Learned normalizing flow mapping on mixture-of-Gaussians and scRNA-seq datasets}
\begin{figure}[!h]
\begin{center}
        \includegraphics[width=\linewidth]{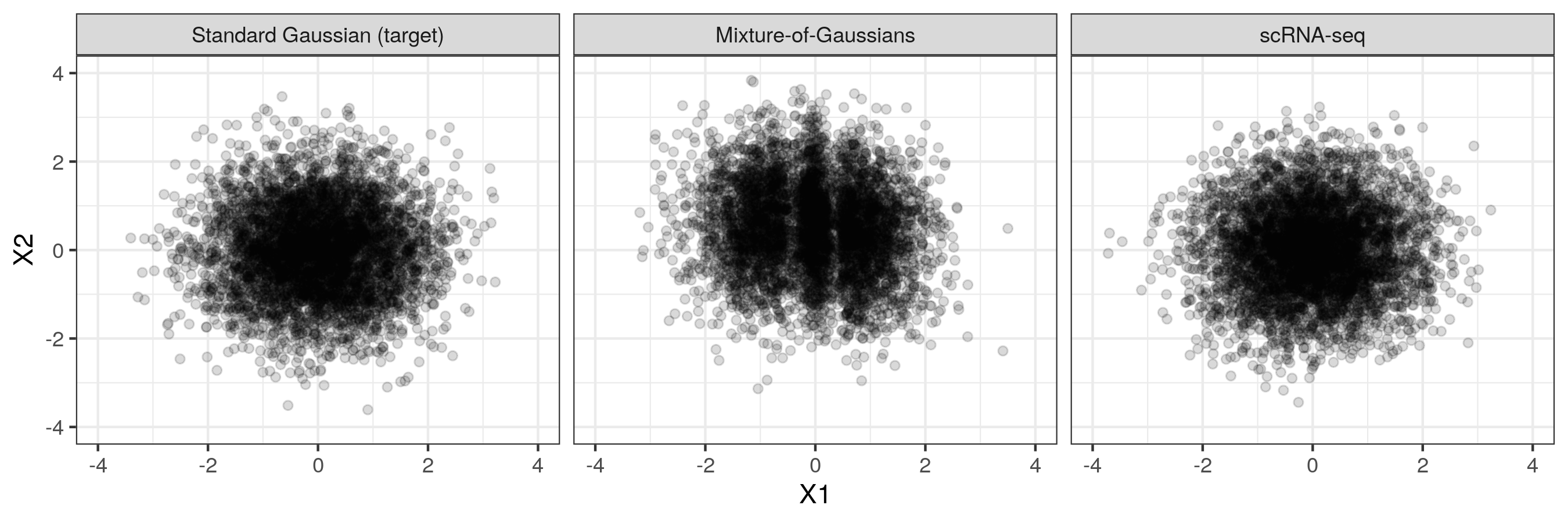}
\end{center}
    \caption{Plot of features mapped to flow space by the learned normalizing flow within \FlowSelect with $j=1$ on the x-axis and $j=2$ on the y-axis. Mapped features are shown for the mixture-of-Gaussians and scRNA-seq datasets, and they are compared to samples from a true standard Gaussian distribution.}
\end{figure}

\end{document}